\documentclass{ieeeaccess}
\usepackage{cite}
\usepackage{amsmath,amssymb,amsfonts}
\usepackage{algorithmic}
\usepackage{graphicx}
\usepackage[dvipsnames]{xcolor}
\usepackage{hyperref}
\usepackage{textcomp}
\usepackage{graphicx}
\usepackage{xcolor}
\usepackage{caption}
\usepackage{stfloats}

\def\BibTeX{{\rm B\kern-.05em{\sc i\kern-.025em b}\kern-.08em
    T\kern-.1667em\lower.7ex\hbox{E}\kern-.125emX}}
\begin{document}
\doi{}

\title{TransNorm: Transformer Provides a Strong Spatial Normalization Mechanism for a Deep Segmentation Model}

\author{\uppercase{Reza Azad}\authorrefmark{1},
\uppercase{MOHAMMAD T. AL-ANTARY\authorrefmark{2}, Moein Heidari}\authorrefmark{3}, \uppercase{Dorit Merhof\authorrefmark{4}
}}

\address[1]{Institute of Imaging and Computer Vision, RWTH Aachen University, Germany. (e-mail: azad@lfb.rwth-aachen.de)}
\address[2]{School of Computing and Mathematical Sciences, University of Greenwich, London SE10 9LS, U.K. (e-mail: m.alantary@gre.ac.uk)}
\address[3]{School of Electrical Engineering, Iran University of Science and Technology, Iran. (e-mail: moein\_heidari@elec.iust.ac.ir)}
\address[4]{Institute of Imaging and Computer Vision, RWTH Aachen University, Germany. (e-mail: dorit.merhof@lfb.rwth-aachen.de)}


\markboth
{Author \headeretal: Preparation of Papers for IEEE TRANSACTIONS and JOURNALS}
{Author \headeretal: Preparation of Papers for IEEE TRANSACTIONS and JOURNALS}

\corresp{Corresponding author: Dorit Merhof (e-mail: dorit.merhof@lfb.rwth-aachen.de).}

\begin{abstract}
In the past few years, convolutional neural networks (CNNs), particularly U-Net, have been the prevailing technique in the medical image processing era. Specifically, the seminal U-Net, as well as its alternatives, have successfully managed to address a wide variety of medical image segmentation tasks.
However, these architectures are intrinsically imperfect as they fail to exhibit long-range interactions and spatial dependencies leading to a severe performance drop in the segmentation of medical images with variable shapes and structures. Transformers, preliminary proposed for sequence-to-sequence prediction, have arisen
as surrogate architectures to precisely model global information assisted by the self-attention mechanism. Despite being feasibly designed, utilizing a pure Transformer for image segmentation purposes can result in limited localization capacity stemming from inadequate low-level features. Thus, a line of research strives to design robust variants of Transformer-based U-Net. In this paper, we propose Trans-Norm, a novel deep segmentation framework which concomitantly consolidates a Transformer module into both encoder and skip-connections of the standard U-Net. We argue that the expedient design of skip-connections can be crucial for accurate segmentation as it can assist feature fusion between the expanding and contracting paths. In this respect, we derive a Spatial Normalization mechanism from the Transformer module to adaptively recalibrate the skip connection path. Extensive experiments across three typical tasks for medical image segmentation demonstrate the effectiveness of TransNorm. The codes and trained models are publicly
available at
\href{https://github.com/rezazad68/transnorm}{\textcolor{red}{github}.}

\end{abstract}


\titlepgskip=-15pt

\maketitle

\section{Introduction}
\PARstart{I}{n} the healthcare field, early diagnosis of diseases is a crucial step as it can aid in detecting the severity and spread of disorders in early stages.
Medical image segmentation plays an integral role in computer-aided disease diagnosis (CAD), treatment planning and surgical pre-assessment. It mainly consists of partitioning the shapes and volumes of target organs and tissues by pixel-wise classification \cite{wu2022d}. Manual annotation in clinical applications is labour-intensive and time-consuming \cite{zhang2021pyramid} and may be prone to human error. Thus, to preclude the burden of manual annotation, automated medical image segmentation has become a direction of research for many years. The recent growth of deep learning in computer vision has prompted researchers to reconsider its potential in CAD. Based on this line of research, deep learning has attained immense success in a wide range of medical domains ranging from abdominal organ segmentation from CT images \cite{azad2019bi, chen2021transunet,heidari2022hiformer}, and skin lesion segmentation from dermoscopy \cite{reza2022contextual} to multiple Mylomia segmentation \cite{bozorgpour2021multi} from microscopic images, MRI images \cite{azad2022medical,azad2021stacked,azad2022intervertebral,azad2022smu} and etc. Figure 1 shows samples of the Synapse multi-organ segmentation dataset \cite{landman2015miccai} which exhibits a diverse set of challenging samples for a medical image segmentation task.

\begin{figure}[h]
\begin{center}
  \includegraphics[width=\linewidth]{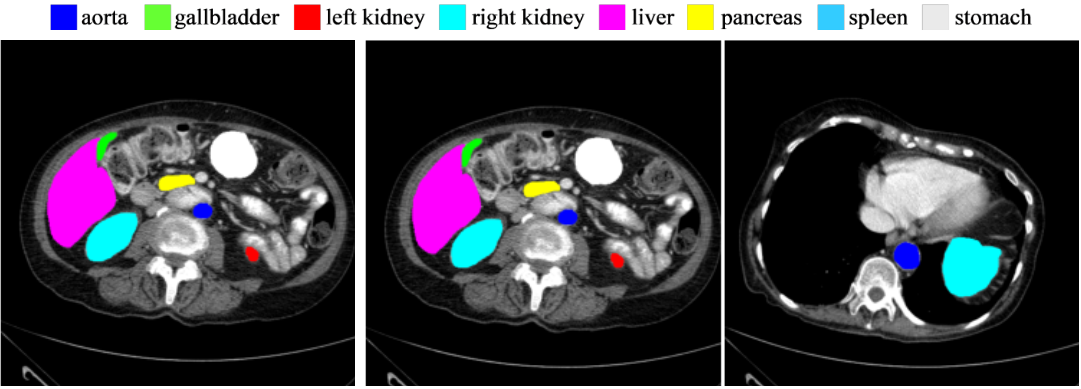}
\end{center}
  \caption{Sample 2D slices of the Synapse multi-organ segmentation dataset and the multi-organ annotation mask.}
\label{fig:figure1}
\end{figure}
Convolutional neural networks (CNNs), and more specifically fully convolutional networks (FCNs) \cite{long2015fully} have been the de facto standard in medical image segmentation for many years. However, these architectures face a major issue, namely that essential details are lost at deeper layers of the network. To solve this problem, and inspired by the seminal U-Net structure \cite{ronneberger2015u}, a family of U-shaped networks were devised. The U-shaped structure, designated as U-Net, and consisting of symmetric encoder-decoder with skip connections has rendered remarkable performances in recent years. The contracting path (encoder) follows the typical structure of a convolutional neural network which encodes the input image into feature representations at multiple levels. The CNN-based models, despite being able to improve context modelling to some degree, have an unavoidable confined receptive field following from the nature of the convolution operation. In natural language processing (NLP), the ubiquitous architecture \emph{Transformer} has been seen as capable of learning long-term features \cite{devlin2018bert,radford2018improving}. Transformers, unlike CNNs, are not only significant at modelling global contexts but are also a promising tool for localizing local details. Inspired by their success in the NLP domain, researchers have inspected leveraging the Transformer network in computer vision. The pioneering architecture, based purely on the self-attention mechanism, was the Vision Transformer (ViT) \cite{dosovitskiy2020image} which attained high performance compared to SOTA in image recognition tasks. Since then, Transformers have proven their prodigious power in diverse computer vision tasks \cite{carion2020end,chen2020generative,reza2022contextual}. To concurrently benefit from the global context modelling of Transformers as well as the capability of CNNs in learning the rich detailed information, many cohort studies have investigated the amalgamation of U-Net and Transformer \cite{chen2021transunet,hatamizadeh2022unetr}. Inspired by this line of research, in this paper, we propose TransNorm, a novel architecture that provides a strong Spatial Normalization mechanism for a deep segmentation model and applies it to medical image segmentation.
Our TransNorm model initially applies the convolution block to encode the input image into a latent space. Next, on top of the generated features, we incorporate the vision Transformer module to capture global contextual dependency from the latent space. The combination of local features attained from the CNN module and the global dependency from the Transformer module provides a complementary representation to enrich the encoder representation. This design is similar to the recently proposed TransUnet approach \cite{chen2021transunet}. 
However, we assert that a feasible design of skip-connections can be crucial for accurate segmentation. Thus, to further enhance the aforementioned design and increase the benefit from the Transformer module we extend this strategy to normalize the skip connection sections of the decoding path. To this end, we define a two-level attention mechanism to adaptively recalibrate the feature combination on the skip connection path. First, by applying the channel attention mechanism, we normalize the feature representation to emphasize the more informative channels. Next, by utilizing the spatial coefficient stemming from the Transformer module, we apply a second level attention mechanism to strengthen the contribution of important regions to the segmentation process. The experimental results on the multi-organ segmentation task show that incorporating a spatial attention module enhances model strength in recovering object boundaries from an overlapped background. 
Our contributions are as follows :

\noindent$\bullet$ Incorporting an attention module on the skip connection

\noindent$\bullet$ Spatial attention module derived from a Transformer model

\noindent$\bullet$ Improved results on the public datasets as well as publicly available implementation source code.

\section{Literature review}
\label{sec:literature}

\subsection{CNN-based Segmentation Networks}
Convolutional neural networks have been the most commonly used approach for image segmentation tasks in recent years. Following the advent of the monumental U-Net \cite{ronneberger2015u}, this elegant design has been incorporated for diverse medical image segmentation tasks. The main contribution of U-Net in this sense is that while upsampling in the network it also concatenates the higher resolution feature maps from the encoder network with the upsampled features in order to more accurately learn representations. Meanwhile, persuaded by the success of U-Net, a line of research attempted to extend this architecture for the purpose of even more accurate segmentation such as Res-UNet \cite{diakogiannis2020resunet}, Dense-UNet \cite{cao2020denseunet}, U-Net++ \cite{zhou2018unet++} and UNet3+ \cite{huang2020unet}, style matching U-Net \cite{azad2022smu,} few-shot U-Net \cite{feyjie2020semi,azad2021texture}. To this end, 3D U-net \cite{cciccek20163d} was proposed as an augmentation of U-Net which deals with 3D volumetric segmentation. Oktay et al. \cite{oktay2018attention} proposed to use \emph{attention gates} in the basic U-Net architecture for pancreas segmentation to enable the network to focus on specific objects of greater importance while ignoring superfluous areas. Furthermore, U-Net++ \cite{zhou2018unet++} proposed to bridge the semantic gap between the feature maps of the encoder and decoder using nested and dense skip connections instead of directly fetching high-resolution feature maps from the encoder into the decoder. The common point of these CNN-based methods is that they inevitably have limitations in capturing long-range dependencies due to the inherited property of the confined receptive field. In fact, the locality and weight-sharing property of convolution operations make them incapable of apprehending global contexts. To date, various methods have been developed to solve this problem of CNN’s restricted receptive field. As such, Yu et al. \cite{yu2015multi} proposed to use atrous convolution with a dilation rate that expands the receptive field without loss of resolution. Zhao et al. \cite{zhao2017pyramid} exploited pyramid pooling at different feature scales so as to agglomerate global information. Wang et al. \cite{wang2018non} proposed a non-local neural network for capturing long-range dependencies by computing the response at a position as a weighted sum of the features at all positions in the input feature maps. Incorporating self-attention modules
into convolutional layers is another attempt to restrain the deficiency of CNNs in non-local modelling capability. To achieve this, Fu et al. \cite{fu2019dual} proposed a dual attention network that models the semantic interdependencies in spatial and channel dimensions using the self-attention mechanism respectively. Despite the efforts made to mitigate the aforementioned problem of CNNs, they still cannot fully meet the clinical application requirements and exacerbate a notorious problem in the segmentation of medical images with various shapes and scales. Specifically, although these methods have been able to boost the performance of segmentation to some extent, the potential of capturing long-range semantic dependencies still needs to be addressed.

\subsection{Transformers} Following the consensus of exploiting Transformer as a de facto operator in the NLP era, more and more Transformer-based methods appear in CV tasks. Specifically, ViT \cite{dosovitskiy2020image} was the primary medium of the Transformer-based methods to surpass the traditional CNN-based architectures in image recognition tasks. Intuitively, ViT divides the input image into multiple partitions (patches), which are then fed into a Transformer encoder followed by an MLP layer to perform the classification. Subsequently, different alternatives of ViT were proposed in the literature such as Swin Transformer \cite{liu2021swin}, LeViT \cite{graham2021levit}, and Twins \cite{chu2021twins}. To solve the drawback of model complexity in regular Transformers, Swin Transformer \cite{liu2021swin} proposes to split image patches into windows, and apply Transformer only within patches inside each window. Additionally, inspired by the basic idea of CNNs, the authors suggest shifting the window and then applying the Transformer module again allowing adjacent windows to interact with each other.

\subsection{Combining CNNs with Transformer architecture}
The pioneering Transformer-based medical image segmentation approaches involve exploiting Transformer layers in the encoder of the architecture. The first
work to do this was the TransUNet paper \cite{chen2021transunet}. As opposed to other ViT-based methods, the authors propose to first convert the input image to a set of low-resolution feature maps by initially inputting the image to a series of convolution layers (using a Res-Net50 backbone) which they then encode with a ViT. The resulting encoded features are then upsampled via so-called Cascaded Upsampler layers in the decoder to output the final segmentation map. To restrain the quadratic complexity of using a purely Transformer-based model which can hinder the performance in segmentation of high-resolution medical images, Cao et al. \cite{cao2021swin} propose Swin-UNet. Inspired by Swin-Transformer they compute the attention within a fixed window. Furthermore, Swin-UNet contains a patch expanding layer for upsampling the decoder’s feature maps thereby reshaping the feature maps of adjacent dimensions into a higher resolution feature map. In another attempt to incorporate the Transformer module into encoding layers,
Wu et al. \cite{wu2022fat} replace the long-established single branch encoder architecture with a dual encoder containing CNNs and a Transformer branch. Moreover, to attain an adaptive feature fusion from these branches to the decoding part, they devise a feature adaptation module (FAM) and a memory-efficient decoder to circumvent the burden of computational inefficiency.

The CNN-based segmentation methods usually suffer from poor global contextual encoding and render a poor prediction of the object boundary. However, the Transformer-based methods are highly capable of encoding the long-range connectively and exhibit a strong feature learning strategy. However, this strategy usually lacks the learning of local information and suffers from poor generalization performance. Thus, combining these two descriptors might provide an efficient feature representation, which is at the heart of our research in this paper. Following the aforementioned TransUnet method (combining CNN and Transformer modules), 
we argue that this method mainly suffers from a weak construction on the skip connection section, hence, we propose an attention module to boost the performance of this pipeline. To this end, we design a two-level attention mechanism based on the Transformer module to adaptively recalibrate the feature combination on the skip connection path. In the next section, we will present our method comprehensively.

\section{Proposed Method}
\begin{figure*}[t]
\centering
  \includegraphics[width=500px]{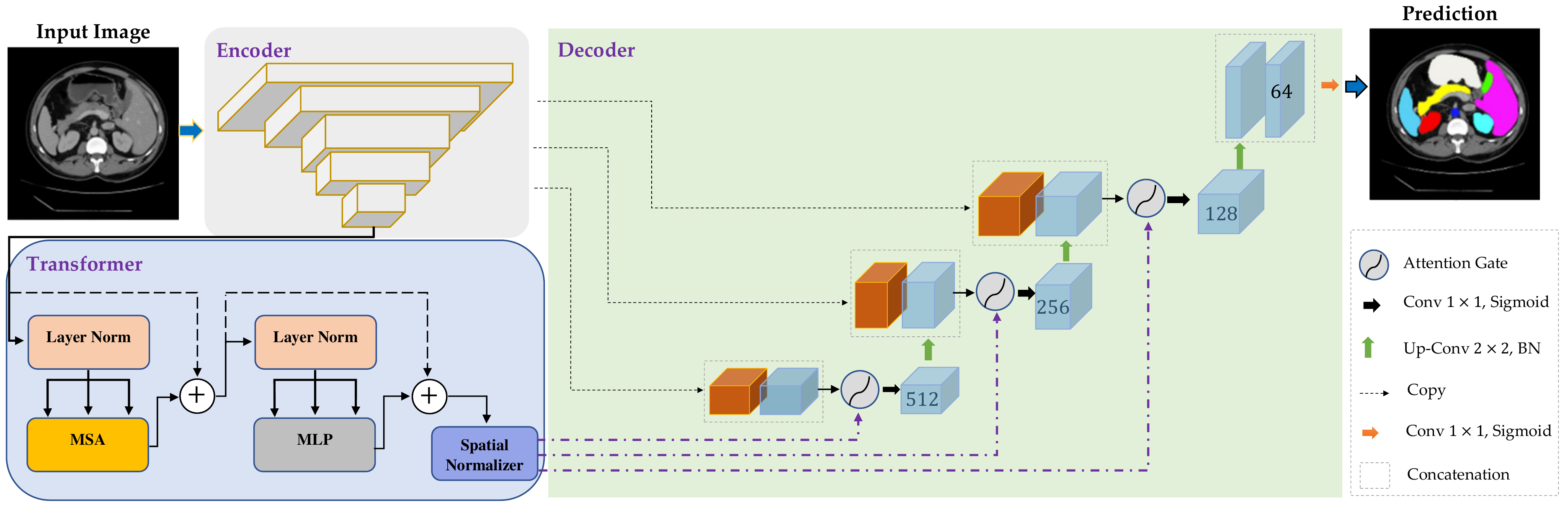}

  \caption{The structure of the proposed method for medical image segmentation. Our approach utilizes a Transformer module on the network bottleneck to learn long-range contextual dependency and produces a spatial normalization coefficient for the attention module.}
\label{fig:figure2}
\end{figure*}

\begin{figure*}[t]
\centering
  \includegraphics[width=500px]{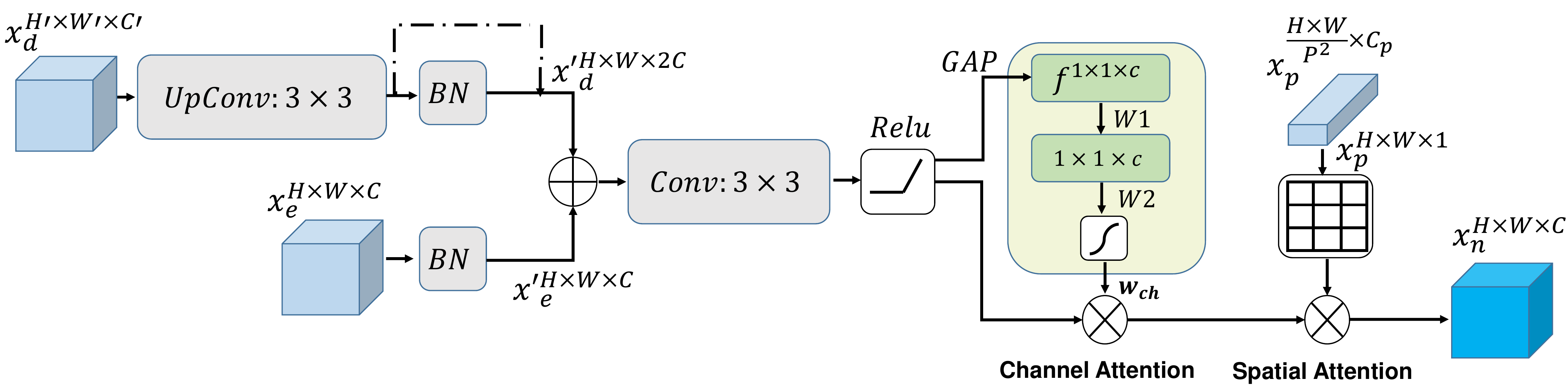}

  \caption{Our proposed two-level attention gate to normalize both channel and spatial information on each decoding path.}
\label{fig:figure3}
\end{figure*}

A conceptual overview of the proposed TransNorm is depicted in Figure \ref{fig:figure2}. Given an image $\mathbf{I} \in \mathbb{R}^{H \times W \times C}$ with a spatial resolution of $H \times W$ and $C$ number of channels, we aim to predict the corresponding segmentation map of size $H \times W$. As shown in Figure \ref{fig:figure2}, TransNorm consists of several components processing information differently: 1) encoder, which is mainly a CNN that encodes semantic and high-level features; 2) Transformer branch, whose goal is to learn the long-range contextual dependency in the bottom stream of our network, within which we propose a so-called \emph{Spatial Normalizer} shown in Figure \ref{fig:figure3}; 3) decoder, to gradually output the final labelmap. We will elaborate each part in more detail in the next subsections.

\subsection{Encoder}
The first module that is incorporated into our design is the encoder module. Generally, the design choice for the encoder module can follow any well-known network, whereas a deeper network may increase the feature representation power and, consequently, offer a better generalization performance. However, the main objective of this study is to investigate the effectiveness and importance of the skip connection in the Transformer-based model. Thus, to squeeze the CNN representation, we exploit the semi U-Net \cite{ronneberger2015u} structure to obtain the semantic and global context of features. Given an input image $\mathbf{I} \in \mathbb{R}^{H \times W \times C}$, the feature maps produced by the encoder $x \in R^{H^{\prime} \times W^{\prime} \times C^{\prime}}$ can be formally expressed as

\begin{equation}
x=E_{\theta}(\mathbf{I}),
\end{equation}
where $E$ denotes the encoder with parameters $\theta$.

\subsection{Long-range Contextual Representation}
To exploit the supremacy of learning long-range correlation and contextual dependency, we integrate the seminal Transformer module following the encoding part of our design. To this end, we perform sequentialization by reshaping the input $x \in R^{H^{\prime} \times W^{\prime} \times C^{\prime}}$ into a sequence of flattened non-overlapping patches $\mathbf{x}_{p} \in \mathcal{R}^{N \times\left(p^{2} \cdot C^{\prime}\right)}$, where each patch is of size $P \times P$ and $N=\frac{H^{\prime} W^{\prime}}{P^{2}}$ denotes the input sequence length. Subsequently, we map the patches into a $D$-dimensional embedding space using a linear projection. To retain the spatial information of each patch, as the attention mechanism is intrinsically permutation-invariant, we learn a position embedding which is later added to the patch embedding as follows:
\begin{equation}
\mathbf{z}_{0}=\left[\mathbf{x}_{p}^{1} \mathbf{E} ; \mathbf{x}_{p}^{2} \mathbf{E} ; \cdots ; \mathbf{x}_{p}^{N} \mathbf{E}\right]+\mathbf{E}_{p o s},
\end{equation}
where $\mathbf{E} \in \mathbb{R}^{\left(P^{2} \cdot C^{\prime}\right) \times D}$ is the patch embedding projection, and $\mathbf{E}_{\text {pos }} \in \mathbb{R}^{N \times D}$ intends to learn the positional encoding. Ultimately, the Transformer encoder is comprised of $K$ stages, each consisting of a Multihead Self-Attention (MSA) and a Multi-Layer Perceptron (MLP) block. Therefore, the output of the $k$-th layer is modeled as:

\begin{equation}
\mathbf{z}_{\emph{k}}^{\prime}=\operatorname{MSA}\left(\mathrm{LN}\left(\mathbf{z}_{\emph{k}-1}\right)\right)+\mathbf{z}_{\emph{k}-1}
\end{equation}
\begin{equation}
\mathbf{z}_{\emph{k}}=\mathrm{MLP}\left(\mathrm{LN}\left(\mathbf{z}_{\emph{k}}^{\prime}\right)\right)+\mathbf{z}_{\emph{k}}^{\prime}
\end{equation}
where $\mathrm{LN}()$ denotes the layer normalization operator \cite{ba2016layer}. We concatenate the generated features with the CNN-based features derived from the encoder to form a complementary (semantic and long-range contextual) feature set. To further benefit from the Transformer module we propose an attention module to provide a Spatial Normalization mechanism for the skip connections in the decoding path.

\subsection{Attention Gate}
\label{section:attention}
The adaptation of attention mechanisms with the aim of diverting focus to salient regions of an image while disregarding irrelevant parts has long been established in the computer vision era. Inspired by its success, we devise a two-level attention mechanism into the skip connection of the decoding path as shown in Figure \ref{fig:figure2}. Given an encoded CNN feature $x \in R^{H^{\prime} \times W^{\prime} \times C^{\prime}}$ and the transformer feature $z \in R^{H^{\prime} \times W^{\prime} \times C^{\prime}}$, our two-level attention module first calculates the 1D channel attention map $\mathbf{W_c} \in \mathbb{R}^{C \times 1 \times 1}$ to adaptively recalibrate the object recognition features and then utilizes a 2D Spatial Normalization map $\mathbf{W_s} \in \mathbb{R}^{1 \times H^{\prime} \times W^{\prime}}$ generated by the Transformer module to perform the Spatial Normalization process. The overall attention process can be written as: 

\begin{equation}
\begin{aligned}
\mathbf{f}_{cn} &=\mathbf{W}_{\mathrm{c}}(\mathbf{f}) \otimes \mathbf{f},\\
\mathbf{f}{sn}&=\mathbf{W}_{\mathbf{s}}\otimes \mathbf{f}_{cn},
\end{aligned}
\end{equation}
Where $\otimes$ shows the element-wise multiplication and $f$ represents the concatenation of the CNN and Transformer features. More specifically, we formulate the channel normalization $f_{cn}$ by adaptively recalibrating the weight of each channel, to feasibly exploit the inter-channel correspondence of feature maps. To this end, we first calculate the global average pooling of each layer. Then, by utilizing a fully-connected layer, we scale each channel accordingly
\begin{equation}
\begin{aligned}
\mathbf{f}_{\mathbf{cn}} &=\sigma(M L P(\operatorname{Avg} \operatorname{Pool}(\mathbf{F}))) \\
&=\sigma\left(\mathbf{W}_{\mathbf{2}}\left(\mathbf{W}_{\mathbf{1}}\left(\mathbf{F}_{\mathbf{a v g}}^{\mathbf{c}}\right)\right)\right),
\end{aligned}
\end{equation}
where $\sigma$ represents the sigmoid activation function, and $W_1$ and $W_2$ show the MLP weights. Next, for the purpose of an adaptive spatial region selection mechanism, we devise a spatial attention map to focus on informative parts of a single feature map. We assert that utilizing the spatial coefficients derived from the Transformer module will lead to a better conclusion for the segmentation model as to where to emphasize or suppress while generating the segmentation map. In this respect, we use the attention probability map $\mathbf{W_s} \in \mathbb{R}^{1 \times H^{\prime} \times W^{\prime}}$ derived from the Transformer module
$
W_s=\operatorname{softmax}\left(\mathbf{q k}^{\top} / \sqrt{d}\right)
$
to perform an attention mechanism:
\begin{equation}
\begin{aligned}
\mathbf{f}{sn}&=\mathbf{W}_{\mathbf{s}}\otimes \mathbf{f}_{cn},
\end{aligned}
\end{equation}
where $q$, $k$, and $d$ represent the query, key and normalization values, respectively. To impose an expedient feature fusion scheme into the decoding part of our proposal, we reconsider the feature recalibration on the skip-connection path. Intuitively, in order to guarantee the aggregation scheme to exploit both the high-level and the with low-level feature maps, we make use of generated attention maps into the skip-connection configuration.

\subsection{Decoder}
The decoder, a pure CNN architecture, progressively upsamples the feature maps to the original image space (\textit{i.e.,} $\mathbb{R}^{H \times W \times C}$).
The cascaded upsampling strategy
is used to recover the resolution from the previous layer using transpose convolution to perform pixel-level segmentation. Moreover, to effectively promote  global information from the Spatial Normalizer module, we utilize the feature maps obtained from the Spatial Normalizer which contain high-level semantic information in the upsample layers using attention gates as shown in Figure \ref{fig:figure2}. We claim that the idea of prominently designing skip-connections leads to finer feature fusion without the severe cost of computational burden. To support our claim, we performed an ablation study on the effect of the afrementioned idea in the following sections.

\section{Experimental Results}
To validate our method for different medical image segmentation tasks, extensive experiments on three different tasks were designed. In this section, first, we briefly present the dataset utilized to evaluate the effectiveness of the designed network. Next, the training hyperparameters and the training process are described in more detail. To validate the performance of the proposed approach compared to the baseline and the counterpart methods, we first describe the metrics and the experimental setting. Then, for each dataset quantitative and visualization results are provided to discuss the obtained results in a comprehensive manner. Furthermore, the ablation study is included to evaluate the contribution and effects of each proposed module separately. In the next subsections, we will discuss each section in more detail.

\subsection{Dataset}

\subsubsection{\textbf{Synapse multi-organ segmentation}}
The Beyond the Cranial Vault (BTCV) abdomen challenge dataset \cite{landman2015miccai} comprises 30 abdominal CT scans with, in total, 3779 axial contrast-enhanced abdominal clinical CT images. In each instance, 13 organs were annotated by interpreters including the spleen, the right kidney, the left kidney, the gallbladder, the esophagus, the liver, the stomach, the aorta, the inferior vena cava (IVC), the portal vein, the splenic vein, the pancreas, the left adrenal gland, and the right adrenal gland. Each CT scan is acquired with contrast enhancement leading to volumes in the range of $85 \sim 198$ slices of $512 \times 512$ pixels, with a voxel spatial resolution of $([0.54 \sim 0.54] \times[0.98 \sim 0.98] \times[2.5 \sim 5.0]) \mathrm{mm}^{3}$. 
Following the splitting strategy of \cite{chen2021transunet}, 18 samples were divided into the training set and 12 samples into the testing set. Moreover, as for the pre-processing pipeline, each scan value was normalized to values between 0 an 1 followed by
random sampling of $96 \times 96 \times 96$ voxels \cite{tang2021self}.

\subsubsection{\textbf{Skin lesion segmentation}}

\textbf{ISIC 2017 Dataset.}
The ISIC 2017 dataset \cite{codella2018skin} was published by the International Skin Imaging Collaboration (ISIC) as a large-scale dataset of dermoscopy images which was primarily rendered for three tasks including lesion segmentation (considered in this work), dermoscopic feature detection, and disease classification. The ISIC 2017 dataset contains a training set of 2000 images with corresponding ground truth images. The size of each sample is 576 × 767 pixels. Following the settings of \cite{azad2021deep}, we use 1250 samples for training, 150 samples for validation data, and 600 samples as a test set. As for the pre-processing stage, we resize images to 256 × 256 pixels following the lead of \cite{alom2018recurrent}.

\textbf{ISIC 2018 Dataset.}
In 2018 the ISIC foundation developed this dataset \cite{codella2019skin} as a large-scale dataset of dermoscopy images with the aim of further supporting clinical research which yields automated algorithmic analysis. The database comprises more than 10,000 dermoscopic images of 7 types of diseases: (melanoma, nevi, seborrheic keratosis, BCC, Bowen’s disease and actinic keratosis, vascular lesions, and dermatofibromas) as well as their corresponding ground truth annotations. We alter the size of each sample i.e., 2016 × 3024, to a fixed size of 256 × 256 pixels.

\textbf{$\mathbf{PH}^{2}$ \text {Dataset}.}
The $\mathrm{PH}^{2}$ dataset \cite{mendoncca2013ph} comprises 200 dermatoscopic images of 
melanocytic skin lesions including 80 common nevi, 80 atypical nevi, and 40 melanomas. The resolution of each input image is 768 × 560 pixels. Following the same strategy as \cite{azad2020attention}, we divide the dataset into train and test sets (100 samples per each set) and resize the images into 256 × 256 pixels. Furthermore, similar to  \cite{azad2020attention} we pre-train the model on ISIC 2017 to later tune it for this dataset.

\subsubsection{\textbf{Multiple Mylomia Segmentation}}
The Multiple Myeloma (MM) challenge was held by the SegPC grand challenge in conjunction with the ISBI 2021 symposium to evaluate the effectiveness of machine learning methods for multiple myeloma cell segmentation \cite{gupta2018pcseg}. This challenge contains several images captured from the bone marrow slides of patients diagnosed with Multiple Myeloma (MM) cancer, which cancer arises from a white blood cell. The annotated train dataset provided by this challenge contains 290 images with an MM instance in each image. To evaluate our method on this dataset, we follow the same strategy presented in \cite{gupta2018pcseg} and divide the original train set into train and validation sets. Our inference is on the original validation set provided by this challenge. 

\subsection{Evaluation Metrics}
We evaluated the performance of Synapse multi-organ segmentation with the average Dice Similarity Coefficient (DSC) and the average Hausdorff Distance (HD). To validate the results obtained on skin lesion segmentation and MM we used several well-known metrics including accuracy (AC), sensitivity (SE), specificity (SP), F1-Score, and mean Intersection over Union (mIoU). The equation of each metric and the brief description is provided as follows: 
TP indicates a sample that is correctly predicted as a true sample; TN shows a sample with a negative label that is correctly predicted as a negative sample; FP indicates that the model predicted a negative sample as a true class while the correct label was the negative class; FN similarly shows a wrong classification for a true positive sample.

\textbf{Accuracy} represents the ratio of the correct classification,

\begin{equation}
A C C=\frac{\mathrm{TP}+\mathrm{TN}}{\mathrm{TP}+\mathrm{TN}+\mathrm{FP}+\mathrm{FN}}
\end{equation}

\textbf{Specificity} calculates the percentage of the FP that are identified by the model,

\begin{equation}
\text { Specificity }=\frac{\mathrm{TN}}{\mathrm{TN}+\mathrm{FP}}
\end{equation}

\textbf{Sensitivity} calculates the percentage of the TP that are identified by the model,

\begin{equation}
\text { Sensitivity/ Recall }=\frac{\mathrm{TP}}{\mathrm{TP}+\mathrm{FN}}
\end{equation}

\textbf{F1 score} calculates the weighted average of the precision vs recall,

\begin{equation}
\text { F1 score }=\frac{2 * \mathrm{TP}}{2 * \mathrm{TP}+\mathrm{FP}+\mathrm{FN}}
\end{equation}

\textbf{mIoU} measures the overlapped area of the predicted mask with the grand truth divided by the union of the grand truth and predicted mask. 

\begin{equation}
m I O U=\frac{T P}{T P+F P+F N}
\end{equation}

\textbf{DSC} measures the similarity of the predicted mask and the grand truth. 

\begin{equation}
D S C=\frac{2 T P}{2 T P+F P+F N}
\end{equation}

\subsection{Implementation details}
To implement the proposed method we used the Pytorch library and carried out the results on a single GPU RTX 3090 equipped System. Both the baseline model and the proposed one are trained with a batch size 16 with an initial learning rate $1e-3$ and the decay rate $1e-4$ for 100 epochs. All layers are initialized using the normal distribution and data augmentation is not used during the training process. To facilitate the training convergence we included the batch normalization layer in each block of the encoding and decoding paths. During the training the validation loss is monitored to stop the training process if the validation loss did not decrease in ten successive epochs. For the baseline model, we followed the common implementation of the U-Net network and incorporated the proposed modules to present our network. It is worthwhile to mention that the training process for all datasets converged smoothly without any instability.

\subsection{Comparison Results}
In this section, the experimental results of each dataset are presented. To provide a fair evaluation we followed the recent work and only included the approaches which use the same settings for their performance evaluation.

\subsubsection{Synapse multi-organ segmentation}
Table 1 presents the comparison results of the proposed method against the SOTA approaches. First, it is clear that our TransNorm achieves the best results, with DSC and HD
of 78.40\% and 30.25\%, respectively. Compared to the recent TransU-Net variants, our method improves the average DSC and Hausdorff distance by $0.98\%$ and 1.44 mm respectively. This may explain the importance of the attention mechanism we incorporated on the skip connections of the decoding path.\\ 
Second, the results presented demonstrate that the TransNorm model improves both CNN and pure Transformer models by a large margin. The CNN models are less capable of capturing the global contextual information and, consequently, less precise in the boundary area. However, the pure Transformer model renders a pool representation of the local information and yields a weak segmentation mask. Combining both these networks and incorporating an attention mechanism, our proposed method results in learning both semantic and global contextual information with an additional adaptation (due to the attention mechanism) which is crucial for the segmentation task.\\

Furthermore, for visual assessment, we have provided Figure \ref{fig:figure1} in which the proposed method's performance is closer to the ground truth, and in line with the real situation. This visualization reveals the importance of the attention mechanism we added to the skip connection module to effectively reconstruct the object boundary on a highly overlapped background.

\begin{figure}[h]
\begin{center}
  \includegraphics[width=\linewidth]{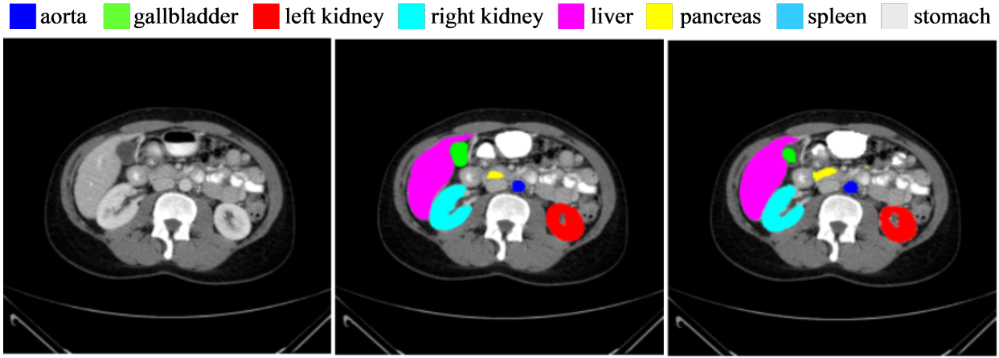}
\end{center}
  \caption{Segmentation result of the proposed method on Synaps dataset. From Left to right, the images indicate the input image, grand truth and the predicted mask.}
\label{fig:figure1}
\end{figure}

\begin{table*}
\centering
\caption{Comparision results of the proposed method on Synaps multi-organ segmentation dataset.}
\resizebox{\textwidth}{!}{
\begin{tabular}{l|cc|cccccccc}
\hline Methods & DSC~$\uparrow$ &HD~$\downarrow$& Aorta & Gallbladder & Kidney(L) & Kidney(R) & Liver & Pancreas & Spleen & Stomach \\

\hline V-Net \cite{milletari2016v} & $68.81$ & $-$ & $75.34$ & $51.87$ & $77.10$ & $80.75$  & $87.84$ & $40.05$&$80.56$& $56.98$\\

R50 U-Net \cite{chen2021transunet} & $74.68$ & $36.87$ &$87.74$&$63.66$&$80.60$&$78.19$&$93.74$&$56.90$&$85.87$&$74.16$
\\

R50 Att-UNet \cite{chen2021transunet} & $75.57$ & $36.97$ &$55.92$&$63.91$&$79.20$&$72.71$&$93.56$&$49.37$&$87.19$&$74.95$
\\

Att-UNet \cite{oktay2018attention} & $77.77$ & $36.02$ &$89.55$&$68.88$&$77.98$&$71.11$&$93.57$&$58.04$&$87.30$&$75.75$
\\

R50 ViT \cite{chen2021transunet}  & $71.29$ & $32.87$ &$73.73$&$55.13$&$75.80$&$72.20$&$91.51$&$45.99$&$81.99$&$73.95$
\\

TransUnet \cite{chen2021transunet} & $77.48$ & $31.69$ &$87.23$&$63.13$&$81.87$&$77.02$&$94.08$&$55.86$&$85.08$&$75.62$
\\
Baseline & $76.85$ & $39.70$ &$89.07$&$69.72$&$77.77$&$68.60$&$93.43$&$53.98$&$86.67$&$75.58$
\\
\hline
\textbf{TransNorm (Proposed)}& $78.40$ & $30.25$ &$86.23$&$65.10$&$82.18$&$78.63$&$94.22$&$55.34$&$89.50$&$76.01$ \\
\hline
\end{tabular}
}
\end{table*}

\subsubsection{Skin Lesion Segmentation} 
In this section, we present the experimental results for skin lesion segmentation. First, in Table 2 we present the comparison results for ISIC 2017. It can be seen that the proposed method outperforms the SOTA approaches in almost all metrics. Compared to the  U-Net model (the best CNN model for this dataset), our method produces a better generalization performance and improves the F1 score by 8$\%$. Furthermore, in comparison with the TransU-Net model, our method exhibits a better performance which further proves the effectiveness of the attention module we incorporated on the skip connection parts. 

\begin{table}[h]
\centering
	\caption{Quantitative analysis of the proposed method against the SOTA approaches for skin lesion segmentation on ISIC 2017.}
	\resizebox{\columnwidth}{!}{
	\begin{tabular}{ccccc}
		\hline
		\textbf{Methods} & \textbf{DSC} & \textbf{SE} & \textbf{SP}&\textbf{ACC} \\
		\hline
         U-Net~\cite{ronneberger2015u} &0.8159 & 0.8172 & 0.9680 & 0.9164\\
         Att U-Net~\cite{oktay2018attention}&0.8082 & 0.7998 & 0.9776 & 0.9145\\
         DAGAN~\cite{lei2020skin}&0.8425 & 0.8363 & 0.9716 & 0.9304\\
         TransUNet~\cite{chen2021transunet}&0.8123&0.8263&0.9577&0.9207\\
    	 MCGU-Net~\cite{asadi2020multi}  &0.8927 & 0.8502 & 0.9855 &  0.9570\\
    	 MedT~\cite{valanarasu2021medical}&0.8037 & 0.8064 & 0.9546 & 0.9090\\
    	 FAT-Net~\cite{wu2022fat}&0.8500 & 0.8392 & 0.9725 & 0.9326\\		
		\hline
		\textbf{Proposed Method} & \textbf{0.8933} & \textbf{0.8532}& \textbf{0.9859} &  \textbf{0.9582}\\
		\hline
	\end{tabular}}
	\label{tab:isic17}
\end{table}

In Figure 5 we illustrate visual output for the method to further deliver a clear view of the smooth segmentation results provided by our approach. 
\begin{figure}[h]
\begin{center}
  \includegraphics[width=\linewidth,height=220px]{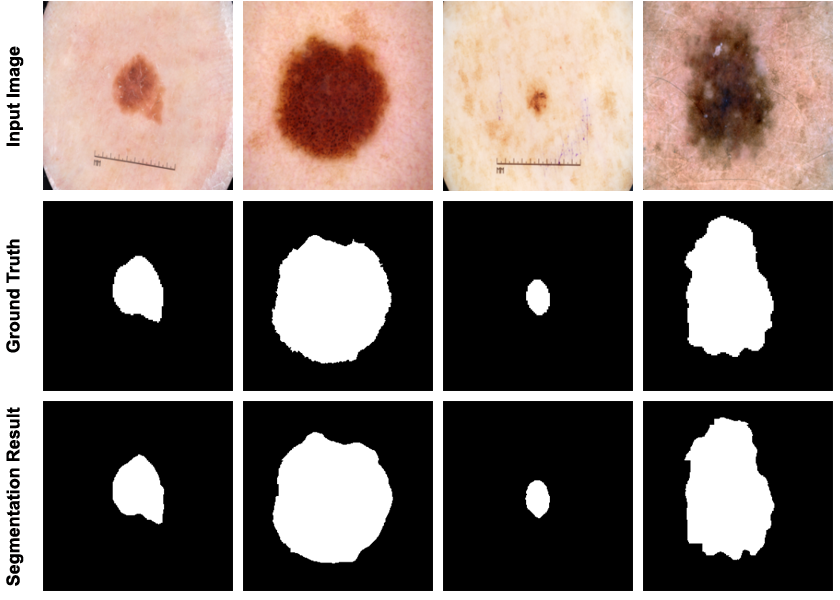}
\end{center}
  \caption{Segmentation results of the proposed method on ISIC 2017. We can observe that the proposed method produces a precise segmentation map and a smooth boundary for the skin lesion area.}
\label{fig:figure6}
\end{figure}

In Table 3 we present the quantitative results of the proposed method against the SOTA approaches. Similar to the ISIC 2018 dataset, our method outperforms both CNN and Transformer-based approaches. To further investigate the qualitative results we depict some segmentation results of the method in Figure 6. The method produces precise segmentation results even when the background has a high overlap with the skin lesion class. 

\begin{table}[h]
\centering
	\caption{Performance comparison of the suggested network against the SOTA counterparts for skin lesion segmentation on the ISIC 2018 dataset.}
	\resizebox{\columnwidth}{!}{
	\begin{tabular}{ccccc}
		\hline
		\textbf{Methods} & \textbf{DSC} & \textbf{SE} & \textbf{SP}&\textbf{ACC} \\
		\hline
         U-Net~\cite{ronneberger2015u} &0.8545 & 0.8800 & 0.9697 &  0.9404  \\
         Att U-Net~\cite{oktay2018attention}&0.8566 & 0.8674 & 0.9863 & 0.9376 \\
         DAGAN~\cite{lei2020skin}&0.8807 & 0.9072 & 0.9588 & 0.9324 \\
         TransUNet~\cite{chen2021transunet}&0.8499 & 0.8578 & 0.9653 & 0.9452\\
    	 MCGU-Net~\cite{asadi2020multi}  &0.895 & 0.848 & 0.986 & 0.955 \\
    	 MedT~\cite{valanarasu2021medical}&0.8389 & 0.8252 & 0.9637 & 0.9358\\
    	 FAT-Net~\cite{wu2022fat}&0.8903 & \textbf{0.9100} & 0.9699 & 0.9578\\	

		\hline
		\textbf{Proposed Method} & \textbf{0.8951} & 0.8750& \textbf{0.9790} &  \textbf{0.9580}\\
		\hline
	\end{tabular}}
	\label{tab:isic18}
\end{table}

\begin{figure}[h]
\begin{center}
  \includegraphics[width=\linewidth,height=220px]{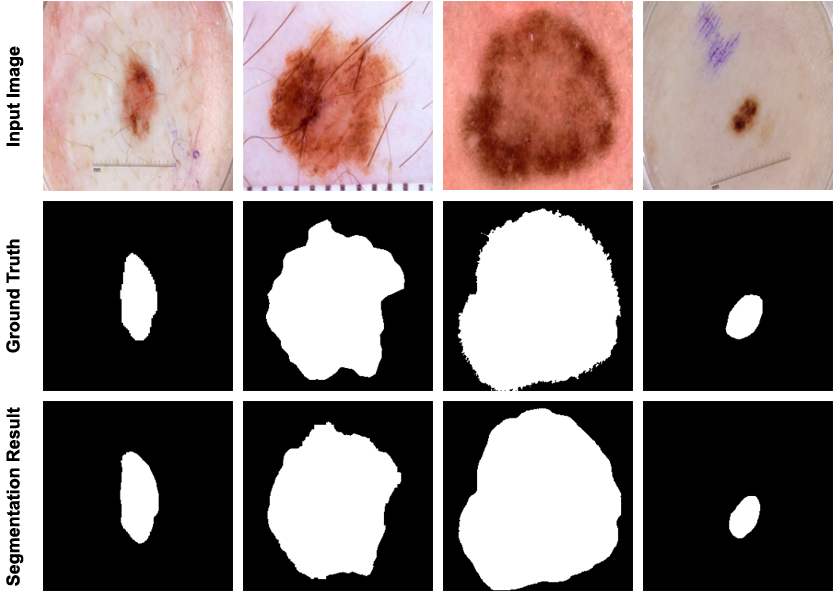}
\end{center}
  \caption{Segmentation samples of the proposed method on ISIC 2018. The method separates the skin lesion from the overlapped background with high precision.}
\label{fig:figure6}
\end{figure}

Finally, Table 4 shows the quantitative results for the PH2 dataset. The table shows that, except for the sensitivity metric, the method exhibits a better performance compared to the SOTA models.

\begin{table}[h]
\centering
    \vspace*{-\baselineskip}
	\caption{Performance comparison on $PH^2$ dataset.}
	\resizebox{\columnwidth}{!}{
	\begin{tabular}{ccccc}
		\hline
		\textbf{Methods} & \textbf{DSC} & \textbf{SE} & \textbf{SP}&\textbf{ACC} \\
		\hline
         U-Net~\cite{ronneberger2015u} &0.8936 & 0.9125 & 0.9588 & 0.9233\\
         Att U-Net~\cite{oktay2018attention}&0.9003 & 0.9205 & 0.9640 & 0.9276\\
         DAGAN~\cite{lei2020skin}&0.9201&0.8320&0.9640&0.9425\\
         TransUNet~\cite{chen2021transunet}&0.8840&0.9063&0.9427&0.9200\\
    	 MCGU-Net~\cite{asadi2020multi}  &0.9263 & 0.8322 & 0.9714  & 0.9537\\
    	 MedT~\cite{valanarasu2021medical}&0.9122 & 0.8472 & 0.9657  & 0.9416\\
    	 FAT-Net~\cite{wu2022fat}&0.9440 & \textbf{0.9441} & 0.9741 & \textbf{0.9703}\\	
		 
		\hline
		\textbf{Proposed Method} & \textbf{0.9437} & 0.9438& \textbf{0.9810} &  \textbf{0.9723}\\
		\hline
	\end{tabular}}
	\label{tab:isic18}
\end{table}

Figure 7 shows segmentation results of the proposed method where the skin lesion area is predicted precisely. Notably, the method does not produce isolated FP pixels and, remarkably, it separates the background from the skin lesion area.  

\begin{figure}[h]
\begin{center}
  \includegraphics[width=\linewidth,height=220px]{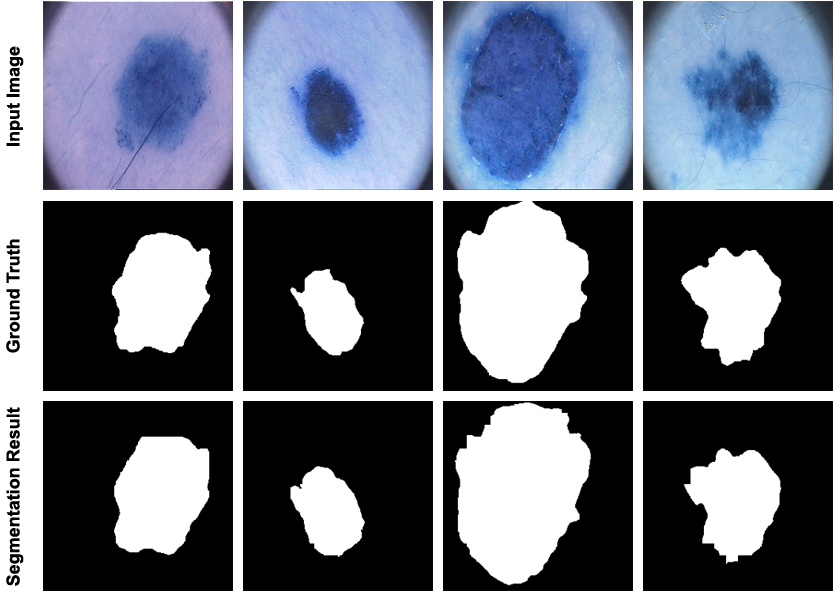}
\end{center}
  \caption{Segmentation samples of the proposed method on PH2. Segmentation results shows that the model is highly capable in separating skin lesion area.}
\label{fig:figure6}
\end{figure}

\subsubsection{Multiple Mylomia Segmentation}
We present the visual segmentation results in Figure 8.One can see in Figure 8 that TransNorm can describe mylomia precisely and generate more fitting segmentation masks. This stems from its ability in modeling long-range spacial correlations between each volume. Furthermore, we conduct quantitative comparisons in terms of the mIOU of the proposed network with the winners of the SegPC 2021 Challenge that utilized the mylomia dataset. As shown in Table 5, the proposed method obtained the best overall
performance. It is worth mentioning that the
leading team (XLAB Insights \cite{bozorgpour2021multi}) propose to use a combination of three instance segmentation networks (SCNet \cite{vu2021scnet},
ResNeSt \cite{zhang2020resnest}, and Mask-RCNN \cite{he2017mask}). The second team (DSC-IITISM \cite{bozorgpour2021multi}) employs the Mask-RCNN model with substantial data augmentation techniques while the third team, (bmdeep \cite{bozorgpour2021multi}) uses an attention deeplabv3+ method \cite{azad2020attention} with a multi-scale region-based training procedure. All these approaches require a large number of parameters compared to our model. However, even with such merit those methods render a poorer performance compared to our novel approach.

\begin{table}
\centering
	\caption{Performance evaluation on the SegPC challenge (best result is highlighted).}
	\resizebox{\columnwidth}{!}{
	\begin{tabular}{cc}
		\hline
		\textbf{Methods} & \textbf{mIOU}\\
		\hline
		Frequency recalibration U-Net \cite{azad2021deep} &0.9392\\		
		XLAB Insights  \cite{bozorgpour2021multi} & 0.9360 \\
		DSC-IITISM  \cite{bozorgpour2021multi} &0.9356\\
		Multi-scale attention deeplabv3+ \cite{bozorgpour2021multi} &0.9065\\
		U-Net \cite{ronneberger2015u} &0.7665\\
		Contexual attention \cite{reza2022contextual}& 0.9395\\
		\textbf{Baseline} & 0.9172 \\
		\hline
		\textbf{Proposed}& \textbf{0.9399}\\
		\hline
	\end{tabular}
	}
	\label{tab:segPC}
\end{table}

\begin{figure}[h]
\begin{center}
  \includegraphics[width=\linewidth,height=220px]{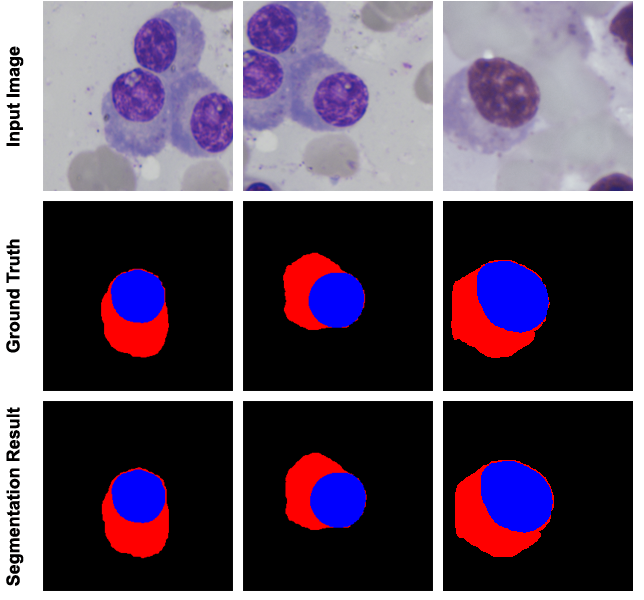}
\end{center}
  \caption{Segmentation results of the proposed method on the SegPC challenge dataset. The method performs well on the highly overlapped background.}
\label{fig:figure6}
\end{figure}

\subsection{Ablation Study}
In this section, we discuss the influence of different design choices on the model's overall performance. To this end, we conducted a comprehensive ablation study on the effect of the attention mechanism, skip connections, input resolution, and model scaling using the ISIC 2018 dataset. In the next subsections, we will elaborate on each factor in detail. 

\subsubsection{Attention Mechanism}
We conducted ablation experiments to demonstrate the values of each contribution and justify the rationale of its design choices towards overall performance. The results, recorded on five metrics are presented in Table 6. Our results reveal the significance of the proposed two-level attention mechanism. In particular, we noticed that integrating both channel and spatial information on the decoding paths would leverage the Transformer to model the global relationships, while exploiting each of them solely lead to less performance boost.

\begin{table}[h]
\centering
	\caption{Performance comparison on ISIC 2018 dataset.}
	\resizebox{\columnwidth}{!}{
	\begin{tabular}{ccc}
		\hline
		\textbf{Methods} & \textbf{DSC}&	\textbf{AC}\\
		\hline
		\textbf{Baseline}&   85.45 & 94.04 \\                 
		\textbf{Baseline+Transformer}&  86.64 & 94.74 \\
		\textbf{Baseline+Transformer+Channel attention}&	0.8771 &	 0.9492 \\	
		\textbf{Baseline+Transformer+Spatial attention}& 0.8821 &  0.9503\\
		\hline
		\textbf{(proposed method)}& \textbf{0.8951} & \textbf{0.9580} \\
		\hline
	\end{tabular}}
	\label{tab:table4}
\end{table}

\subsubsection{Skip Connection Influence}
The objective of the skip connection in our model is to enhance the segmentation performance by providing a low-level spatial information for the decoding path. We argue that the number of skip connections incorporated in our design can remarkably effect the model performance. To analyze such influence, we have conducted several experiments using a varying number of skip connections in the model structure. Table 7 shows the experimental results. It can be seen that the model with three skip connections in 1/4, 1/8, and 1/16 resolution scales provides a better performance compared to the other settings.


\begin{table}
\centering
	\caption{The influence of the number of skip connections on the model performance. Experiments on the ISIC 2018 dataset.}
	\resizebox{\columnwidth}{!}{
	\begin{tabular}{ccc}
		\hline
		\textbf{~ ~ ~Skip Connection count ~ ~ ~} & \textbf{DSC}&\textbf{AC}\\
		\hline
		0 &   0.8635 & 0.9440 \\                 
		1 &  0.8714 & 0.9494 \\
		2 &	0.8861 &  0.9522 \\	
		\textbf{3}& \textbf{0.8951} & \textbf{0.9580} \\
		\hline
	\end{tabular}}
	\label{tab:table4}
\end{table}

\subsubsection{Input Resolution Influence}
The input resolutions utilized in our experiments for the skin lesion segmentation follows the literature work \cite{azad2021deep,lei2020skin} and uses the input resolution 256*256 pixels. However, the high-resolution input ($512 \times 512$) samples can effectively influence the segmentation results as they provide a finer details of the object of interest. In Table 8 we conducted an experiment using higher image resolution to demonstrate the impact of this factor on the model performance. Although the higher resolution provides a better DSC score and justifies the model capability in learning a finer segmentation map, it largely increases the computation complexity. Hence, the performance of our model could be further increased, but this would demand extra computational cost.

\begin{table}[h]
\centering
	\caption{The influence of the input resolution on the model performance. Experiments on ISIC 2018 dataset.}
	\resizebox{\columnwidth}{!}{
	\begin{tabular}{ccc}
		\hline
		\textbf{~ ~ ~ ~Input Resolution size ~ ~ ~} & \textbf{DSC}&	\textbf{AC}\\
		\hline
		\textbf{$256 \times 256$}&   0.8951 & 0.9580 \\                 
		\textbf{($512 \times 512$)}& \textbf{0.9131} & \textbf{0.9650} \\
		\hline
	\end{tabular}}
	\label{tab:table4}
\end{table}

\subsubsection{Model Scaling Influence}
In another setting, we evaluated the effect of network deepeing on the overall performance. To this end, a larger version of the model with a higher number of parameters is created to gain a better performance. However, the experimental results (Table 9) demonstrated that increasing the number of parameters only provides a better performance gain for the training set (overfitting) and, consequently, little improvement is achieved at inference time. Thus, model scaling is not efficient in terms of the computational burden for our design. 

\begin{table}[h]
\centering
	\caption{The influence of the input resolution on the model performance. Experiments on the ISIC 2018 dataset.}
	\resizebox{\columnwidth}{!}{
	\begin{tabular}{ccc}
		\hline
		\textbf{~ ~ ~ ~ Model Scale selection ~ ~ ~ ~} & \textbf{DSC}&	\textbf{AC}\\
		\hline
		Base &   0.8951 & 0.9580 \\                 
		\textbf{Larger version}& \textbf{0.8963} & \textbf{0.9588} \\
		\hline
	\end{tabular}}
	\label{tab:table4}
\end{table}

\section{Conclusion}
In this paper, we presented our Transnorm model for the medical image segmentation task. Our design uses the benefit of combining Transformer and CNN features to encode both semantic and long-range contextual features.  It further utilizes an attention mechanism on the skip connection sections to adaptively recalibrate the feature representation power and, thus, boost the generalization performance. Experimental results on several datasets demonstrate the effectiveness of our approach for smooth and precise segmentation results which is crucial for a clinical application.

\bibliographystyle{unsrt}
\bibliography{Ref}

\EOD
\end{document}